%% file: main.tex
\documentclass[sigconf]{acmart}

\AtBeginDocument{%
  \providecommand\BibTeX{{%
    \normalfont B\kern-0.5em{\scshape i\kern-0.25em b}\kern-0.8em\TeX}}}

\setcopyright{acmcopyright}
\copyrightyear{2023}
\acmYear{2023}
\acmDOI{XXXXXXX.XXXXXXX}

\acmConference[Conference acronym 'XX]{Make sure to enter the correct
  conference title from your rights confirmation email}{June 03--05,
  2018}{Woodstock, NY}
\acmPrice{15.00}
\acmISBN{978-1-4503-XXXX-X/18/06}

\acmSubmissionID{0493}

\usepackage{microtype}
\usepackage{mathtools}

\usepackage{amsthm}

\usepackage{subfigure}  
\usepackage{algorithmicx}  
\usepackage{algpseudocode}  
\usepackage{amsmath}
\usepackage{xspace}

\usepackage{xcolor}

\makeatletter
\DeclareRobustCommand\onedot{\futurelet\@let@token\@onedot}
\def\@onedot{\ifx\@let@token.\else.\null\fi\xspace}

\def\ie{\emph{i.e}\onedot}

\def\etal{\emph{et al}\onedot}

\begin{document}

\newcommand{\name}[0]{MNA-GT\xspace}
\newcommand{\pname}[0]{MNA\xspace}
\title{Adaptive Multi-Neighborhood Attention based Transformer for Graph Representation Learning}


\author{Gaichao Li$^{1,2,\ast}$, Jinsong Chen$^{1,2,\ast}$, Kun He$^{1,\dag}$}

\thanks{$\ast$ Both authors contributed equally to this research.}
\thanks{$\dag$ Corresponding author.}
\affiliation{%
  \institution{$^{1}$ School of Computer Science and Technology, Huazhong University of Science and Technology   
  \city{Wuhan}
  \country{China}}
  \institution{$^{2}$ Institute of Artificial Intelligence, Huazhong University of Science and Technology
  \city{Wuhan}
  \country{China}}
}
\email{{gaichaolee, chenjinsong, brooklet60}@hust.edu.cn}

\def\authors{Gaichao Li, Jinsong Chen, Kun He}

\renewcommand{\shortauthors}{Gaichao, et al.}

\begin{abstract}
By incorporating the graph structural information into Transformers, graph Transformers have exhibited promising performance for graph representation learning in recent years.
Existing graph Transformers leverage specific strategies, such as Laplacian eigenvectors and shortest paths of the node pairs, to preserve the structural features of nodes and feed them into the vanilla Transformer to learn the representations of nodes. 
It is hard for such predefined rules to extract informative graph structural features for arbitrary graphs whose topology structure varies greatly, limiting the learning capacity of the models. 
To this end, we propose an adaptive graph Transformer, termed  Multi-Neighborhood Attention based Graph Transformer (\name), which captures the graph structural information for each node from the multi-neighborhood attention mechanism adaptively. 
By defining the input to perform scaled-dot product as an attention kernel, \name constructs multiple attention kernels based on different hops of neighborhoods such that each attention kernel can capture specific graph structural information of the corresponding neighborhood for each node pair.
In this way, \name can preserve the graph structural information efficiently by incorporating node representations learned by different attention kernels. 
\name further employs an attention layer to learn the importance of different attention kernels to enable the model to adaptively capture the graph structural information for different nodes. 
Extensive experiments are conducted on a variety of graph benchmarks, and the empirical results show that \name outperforms many strong baselines.

\end{abstract}



\maketitle

\section{Introduction}
\input{01_Intro.tex}
\section{Related Work} \label{RW}
\input{05_RW.tex}

\section{Preliminaries}
\input{02_Pre.tex}

\section{Methodology}
\input{03_Method.tex}

\section{Experiments}
\input{04_Exp.tex}

\section{Conclusion}
\input{06_Con.tex}

\begin{acks}
To Robert, for the bagels and explaining CMYK and color spaces.
\end{acks}

\bibliographystyle{ACM-Reference-Format}
\bibliography{WWW2023/07_Reference}



\end{document}

%% file: 01_Intro.tex
 
Over the last decade, benefiting from the massage passing mechanism~\cite{mpnn} that aggregates the features of immediate neighbors iteratively, Graph Neural Networks (GNNs)~\cite{gcn,gat,gdc} have shown promising performance on various graph learning tasks and have been recognized as 
one of the leading models for graph representation learning.

However, there is a growing recognition that the massage passing mechanism of GNNs also cause some inherent limitations, including problems of over-smoothing~\cite{deepgcn1} and over-squashing~\cite{oversq}, thus restricting the expressiveness capability of the models.
There is an urgent requirement to develop new architectures to overcome the 
limitations of massage passing-based GNNs and achieve advanced capability for graph representation learning.

On the other hand, as a powerful and general model derived from the self-attention mechanism, Transformer~\cite{transformer} has received great success on a broad class of machine learning tasks in various domains, such as natural language processing~\cite{trans-survey} and computer vision~\cite{swintransformer}.
Consequently, graph Transformer~\cite{graphtrans,SAN,gt,sat,graphormer} emerges that adopts the Transformer architecture into graph representation learning. 

The main challenge of using Transformer to model graph structural data is that the original Transformer is unable to capture the graph structural information since the self-attention mechanism directly calculate the interaction information for each pair of nodes, which is equivalent to regarding all node as fully connected.
To fill this gap, existing graph Transformers develop various strategies to incorporate the structural information of graphs into the Transformer architecture to learn more powerful node representations.
The strategies can be mainly divided into two categories:
(1) Strengthening the input of Transformer.
Several methods combine the Laplacian eigenvectors~\cite{gt,SAN} with the node attribute feature vectors or the hidden vectors learned by a specific GNN~\cite{graphtrans,sat} with fixed layers as the inputs of Transformer for preserving both the structural and attribute information of nodes.
(2) Introducing the graph structural bias into the attention matrix.
Another line of graph Transformers computes the structural bias of node pairs, such as the shortest path~\cite{graphormer}, and further injects them into the attention matrix to enable the Transformer to capture graph structural information.
By incorporating various graph structural information into the Transformer architecture, graph Transformers have surpassed GNNs on a series of graph mining tasks, such as graph classification and graph regression.

Despite effectiveness, 
existing graph Transformers suffer from leveraging specific strategies to extract graph structural information.
For various nodes, these models uniformly capture the structural information by utilizing the Laplacian eigenvectors derived from fixed number of eigenvalues or the GNN model with fixed number of layers.
However, different graphs exhibit different topological structures, reflecting diverse structural information.
Such fixed strategies can not adaptively capture structural information for various nodes from different graphs, severely limiting the expressiveness of graph Transformer for graph representation learning.


To this end, we propose an adaptive graph Transformer, called Multi-Neighborhood Attention based Graph Transformer (\name).
Different from existing graph Transformers that utilize fixed form of methods to capture the structural information, \name can adaptively obtain the structural information for each node with the novel multi-neighborhood attention (\pname) module. Recalling that each attention kernel represents a triplet to perform self-attention, \pname first constructs multiple attention kernels based on different neighborhoods. With each kernel learning the structural information of the corresponding neighborhood, for each node, \pname can thus generate multiple node representations that inject different neighborhood information separately. Then, \pname further leverages an attention layer to measure which attention kernel matters in the node level so as to adaptively capture the graph structural information for various nodes. To validate the effectiveness of \name, we conduct experiments on five widely used graph datasets for the graph classification task. The empirical results show that our proposed model achieves significant performance improvement comparing with many state-of-the-art graph Transformers and other mainstream models. 

The main contributions of our work are summarized as follows: 
\begin{itemize}
    \item We propose a novel graph Transformer named \name that can perform self-attention on multiple neighborhoods simultaneously to capture more affluent structural information of the graphs.
    \item We further develop an attention module for \name to adaptively incorporate graph structural information for various nodes.
    \item Extensive experiments on public graph benchmarks demonstrate the superiority of our proposed \name over many advanced baselines.  
\end{itemize}

%% file: 05_RW.tex
This section briefly reviews recent works of graph Transformers.

Since the standard Transformer only encodes node attribute information, 
existing graph Transformers aim to introduce graph structural information into the Transformer architecture so as to learn the node representations from both the attribute information and graph structural information. 
Various strategies have been developed to achieve this goal, which can be divided into the following two categories.


\textbf{Strengthening the Transformer Input.}
Methods of this category attempt to strengthen the node attribute features with graph structural information as the input to the Transformer. 
Some researchers extract the structural information directly from the adjacency matrix.
Dwivedi~\etal~\cite{gt} pre-calculate the Laplacian eigenvectors and fuse them with the attribute feature vectors via concatenation. 
Similarly, Kreuzer~\etal~\cite{SAN} utilize the full Laplacian spectrum to learn the structural feature vectors and add them to the attribute features which are then utilized to train the Transformer model. 
Park~\etal~\cite{grpe} propose a relative positional encoding called Graph Relative Positional Encoding (GRPE), which can encode both node-spatial relation and node-edge relation with two sets of learnable positional encoding vectors.
Meanwhile, other researchers focus on combining graph neural network (GNN) with Transformer. By directly stacking the GNN layers before a standard Transformer architecture, GraphTrans~\cite{graphtrans} utilizes GNN to extract local structural information and further feed them into the Transformer to obtain long-range pairwise relationships. To learn structure-aware attention, SAT~\cite{sat} extracts a subgraph representation rooted at each node and further identifies structural similarity between the nodes. 
Grover~\cite{grover} employs two GTransformer modules to represent the node-level and edge-level features, respectively. The inputs of each GTransformer are first fed into a tailored GNN called dyMPN to capture the local information.
However, as mentioned above, these two kinds of methods suffer from a common drawback that they can not encode diverse structural information via a fixed strategy for various nodes.

\textbf{Encoding Graph Structural Bias into the Attention.}
Another line of works~\cite{gt,graphormer} encodes a bias term in the attention matrix to capture the spatial dependency in a graph. 
The bias is derived from the graph structural information to enhance the capability of Transformer to capture graph structural features. 
GT~\cite{gt} applies an attention mask mechanism with the adjacency matrix, restricting each node to only attend to the local neighbors in the graph. 
Mialon~\etal~\cite{graphit} further propose GraphiT that extends the adjacency matrix to a kernel matrix, which is more powerful and flexible to encode various graph kernels.
Graphormer~\cite{graphormer} adopts the degree centrality and shortest path distance between node pairs to capture their structural similarity, which can be further regarded as the bias being encoded to the attention matrix to improve the effectiveness of the Transformer. 
Gophormer~\cite{gophormer} encodes $\mathrm{M}$ views of structural information of each node pair as a proximity encoding vector to enhance the corresponding attention score. 
Nevertheless, introducing graph structural bias obtained by a subjective and certain way is insufficient to measure different structural connections from different graphs.

In this work, we propose \name that aims at adaptively capturing structural information for various nodes, rather than utilizing a fixed strategy.

%% file: 02_Pre.tex
\begin{figure*}
  \includegraphics[width=15cm]{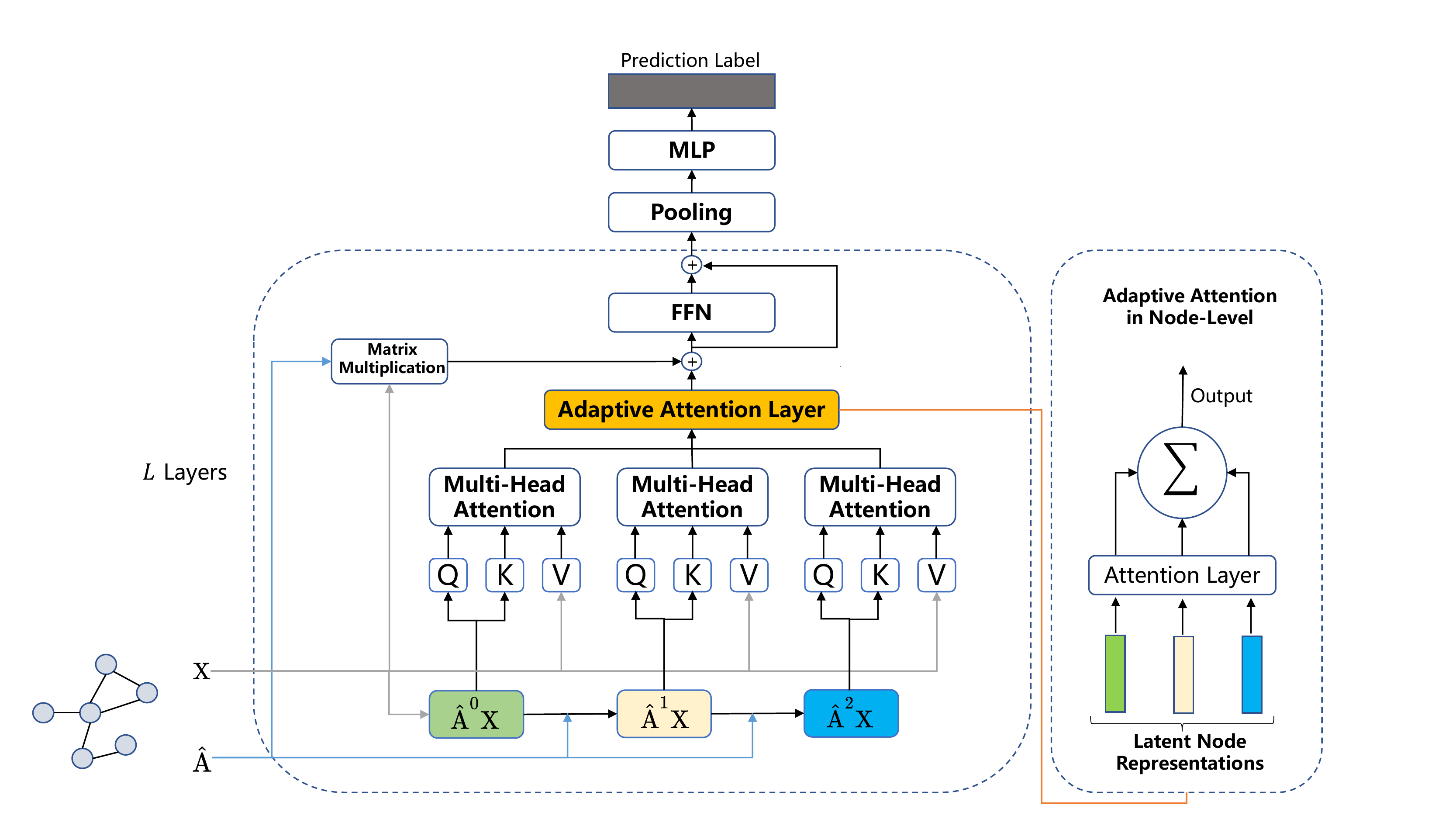}
  \caption{The architecture of \name with three attention kernels. \name first constructs three attention kernels based on the corresponding  neighborhoods. With each attention kernel independently performing multi-head attention, \name can efficiently preserve different structural information encoded by different neighborhoods. \name further utilizes an attention layer to adaptively aggregate the hidden representations for various nodes. Finally, a pooling layer and a multi-layer perceptron are adopted for downstream graph classification tasks.}
  \label{fig:frame}
\end{figure*}

For preliminaries, we first present the notations and terminologies of graph data for problem formulation, then recap the classic Transformer architecture. 

\subsection{Notation}
We define an unweighted and undirected graph as $\mathcal{G}=(\mathcal{V}, \mathcal{E})$, with $\mathcal{V}=\{v_1, v_2, ..., v_n\}$ being the set of nodes and $n=|\mathcal{V}|$ being the number of nodes. For each node $v_i \in \mathcal{V}$, $\mathbf{x}_i \in \mathbf{X}$ is the corresponding feature vector, where 
$\mathbf{X} \in \mathbb{R}^{n \times d}$ is the feature matrix of $\mathcal{G}$ and $d$ is the dimension of feature vector. Let 
$\mathbf{A} \in \mathbb{R}^{n \times n}$ denote the adjacency matrix, where $\mathbf{A}_{ij}=1$ if there exists an edge between the node $v_i$ and $v_j$. Besides, $\tilde{\mathbf{A}} = \mathbf{A} + \mathbf{I}_n$ denotes the adjacency matrix with added self-loops. We can further obtain the normalized adjacency matrix $\hat{\mathbf{A}}$ by symmetrical normalization $\hat{\mathbf{A}}=\mathbf{D}^{-\frac{1}{2}} \tilde{\mathbf{A}} \mathbf{D}^{-\frac{1}{2}}$ or random walk normalization $\hat{\mathbf{A}} = \mathbf{D}^{-1} \tilde{\mathbf{A}}$, where $\mathbf{D}$ is the diagonal degree matrix derived from $\tilde{\mathbf{A}}$. 

\subsection{Transformer}
The Transformer architecture~\citep{transformer} is originally designed for processing machine translation tasks, which is solely based on the attention mechanism. Here we only illustrate the Transformer encoder due to different downstream tasks. An encoder layer is composed of two key components: a multi-head attention module (MHA) and a position-wise feed-forward network (FFN). The MHA module, which stacks several scaled dot-product attention layers, is focused on calculating the similarity between 
queries and keys. Specifically, let $\mathbf{H} \in \mathbb{R}^{n \times d}$ be the input of self-attention module where $n$ is the number of input tokens and $d$ is the hidden dimension. Then the input is first projected through linear transformations to obtain queries $\mathbf{Q}$, keys $\mathbf{K}$ and values $\mathbf{V}$:
\begin{equation}
	\mathbf{Q} = \mathbf{HW}_Q,~
	\mathbf{K} = \mathbf{HW}_K,~
	\mathbf{V} = \mathbf{HW}_V,
	\label{eq:QKV} 
\end{equation}
where $\mathbf{W}_Q \in \mathbb{R}^{d \times d_Q}, \mathbf{W}_K \in \mathbb{R}^{d \times d_K}, \mathbf{W}_V \in \mathbb{R}^{d \times d_V}$ are trainable projection matrices. The self-attention can be further calculated as:
\begin{equation}
    \mathrm{Attention}(\mathbf{Q}, \mathbf{K}, \mathbf{V}) = \mathrm{softmax}(\frac{\mathbf{QK}^{\top}}{\sqrt{d_{out}}})\mathbf{V},
    \label{eq:Attention}
\end{equation}
where $d_{out}$ refers to the dimension of $\mathbf{Q}$. In practice, we usually perform multi-head attention by running several attention layers in parallel which allows the model to jointly attend to information from different subspaces at different positions. By concatenating multiple instances of \autoref{eq:Attention}, the multi-head attention with $m$ heads can be expressed as:
\begin{equation}
    \mathrm{MHA}(\mathbf{H}) = \mathrm{Concat}(\mathrm{head}_1,...,\mathrm{head}_m)\mathbf{W}_O,
    \label{eq:MHSA}
\end{equation}
where $\mathrm{head}_j = \mathrm{Attention}(\mathbf{Q}_j, \mathbf{K}_j, \mathbf{V}_j)$ and $\mathbf{W}_O \in \mathbb{R}^{md_V \times d}$ is a parameter matrix. 
$\mathrm{Concat}(\cdot)$ denote the concatenation operation.
The output of MHA is followed by a skip-connection, and then fed into a FFN module consisting of two linear transformations with a ReLU activation in between.

%% file: 03_Method.tex
In this section, we illustrate our proposed adaptive Multi-Neighborhood Attention based Transformer (\name) in detail. 
The overall framework of \name is depicted in Figure \ref{fig:frame}. We first elaborate its novel attention mechanism module, which can adaptively capture the structural information between nodes from different hops of neighborhoods, thus is more suitable to graph-structured data. Then, we introduce the implementation details of a \name layer. Finally, we compare our proposed \name with several well-known graph Transformers to investigate its expressive ability. 


\subsection{Multi-Neighborhood Attention}
Here we concretely introduce the proposed Multi-Neighborhood Attention in this subsection.

\textbf{Attention Kernel. }
For convenience, 
we first define the input to perform scaled dot-production as a triplet $(\mathbf{H}_Q, \mathbf{H}_K, \mathbf{H}_V)$, namely a group of attention kernel. Based on this definition, \autoref{eq:QKV} and \autoref{eq:Attention} can be reformulated as:
\begin{equation}
    \mathrm{Attention}(\mathbf{Q}, \mathbf{K}, \mathbf{V}) = \mathrm{Attention}(\mathbf{H}_Q\mathbf{W}_Q, \mathbf{H}_K\mathbf{W}_K, \mathbf{H}_V\mathbf{W}_V).
    \label{eq:pk}
\end{equation}
Obviously, when applying to graphs, the attention kernel of the original Transformer~\citep{transformer} is $(\mathbf{X}, \mathbf{X}, \mathbf{X})$,  which only catches the attributed similarity between nodes. To take the structural information into account, SAT~\citep{sat} leverages $(\mathrm{GNN}(\mathbf{X}), \mathrm{GNN}(\mathbf{X}), \mathbf{X})$ as its attention kernel, where $\mathrm{GNN}(\cdot)$ is a structure extractor that can compute subgraph representations centered at each node. Positional encoding based methods~\citep{gt,grpe,SAN} can be considered as taking $(\mathbf{X+PE}, \mathbf{X+PE}, \mathbf{X+PE})$ as an attention kernel, where $\mathbf{PE}$ indicates the positional encoding matrix of all nodes. 
However, the structural information of different graphs varies in the real world. Furthermore, the most mattered structural information are different across different graphs. Thus, by encoding structural information via fixed rules, these methods are not able to capture the structural information of different graphs adaptively. In contrast to these models, our proposed \name first constructs multiple groups of attention kernels based on different hops of neighborhoods, which can capture different kinds of structural information parallelly. Then \name adaptively aggregates the outputs of attention kernels that reflect diverse structural information.  

\textbf{Multi-Neighborhood Attention Mechanism. }
We construct multiple attention kernels based on different hops of neighborhoods to encode the structural information.
Let $\mathcal{N}^k(v_i)$ be the $k$-hop neighborhood set of a node $v_i$ and $\mathbf{x}_i$ be its feature vector. 
We define $\mathcal{N}^0(v_i)=v_i$, which means the 0-hop neighborhood of $v_i$ is itself.
With an aggregation operator $\phi$, $v_i$ can aggregate information from its $k$-hop neighborhood and update itself as $\mathbf{x}_i^k$: 
\begin{equation}
    \mathbf{x}_i^k = \phi(\mathcal{N}^k(v_i)),
    \label{eq:k_agg}
\end{equation}
where 
$k \in \{0,1,2..., c\}$, and $c$ is a hyperparameter. 
Herein, we can obtain $c+1$ neighborhood representations of $v_i$ with each hop corresponding to a representation. The $k$-hop similarity between a pair of nodes $v_i$ and $v_j$ can be further formulated as: 
\begin{equation}
    Sim_{ij}^k = f(\mathbf{x}_i^k, \mathbf{x}_j^k),
    \label{eq:sim}
\end{equation}
where $f(\cdot)$ is a function to calculate similarity and $Sim_{ij}^k$ represents the $k$-hop similarity between $v_i$ and $v_j$. Since $k$ ranges from $0$ to $c$, the structural similarity between $v_i$ and $v_j$ can be reflected from $c+1$ views. 

Extending to the whole feature matrix $\mathbf{X}$ for all nodes, we utilize the normalized adjacency matrix $\hat{\mathbf{A}}$ as the aggregation operator. Similar to the propagation process in~\citet{gdc}, the $k$-hop neighborhood matrix can be denoted by $\hat{\mathbf{A}}^k\mathbf{X}$. Based on these neighborhood matrices, we can develop a variety of attention kernels:
\begin{equation}
    \mathrm{Att-Kernel} = \{(\hat{\mathbf{A}}^k\mathbf{X}, \hat{\mathbf{A}}^k\mathbf{X}, \mathbf{X}) \mid k\in \{0,1,2..., c\}\}.
    \label{eq:pk_0hop}
\end{equation}

For instance, when applying the kernel $(\hat{\mathbf{A}}^2\mathbf{X}, \hat{\mathbf{A}}^2\mathbf{X}, \mathbf{X})$ to perform scaled-dot attention, it means to aggregate the original node representation in a global way according to the weights calculated by 2-hop neighborhood similarity. 
We can also leverage GNN based models to construct different attention kernels by stacking multiple layers, with each kernel corresponding to the output of a layer. 

Similar to the original Transformer\citep{transformer}, our \name applies multi-head attention mechanism for each attention kernel to learn more affluent information from different subspaces. Taking the kernel $(\hat{\mathbf{A}}^k\mathbf{X}, \hat{\mathbf{A}}^k\mathbf{X}, \mathbf{X})$ as an example, we first linearly project $\hat{\mathbf{A}}^k\mathbf{X}$, $\hat{\mathbf{A}}^k\mathbf{X}$ and $\mathbf{X}$ with different learnable matrices for $m$ times (\ie, $m$ heads), respectively. Then the outputs of $m$ heads are concatenated and once again projected, resulting in the output of this kernel:
\begin{equation}
    \mathbf{Z}^k = \mathrm{MHA}(\hat{\mathbf{A}}^k\mathbf{X}, \hat{\mathbf{A}}^k
    \mathbf{X}, \mathbf{X}) = \mathrm{Concat}(\mathrm{head}_1,...,\mathrm{head}_m)\mathbf{W}_O,  
\end{equation}
where $\mathrm{head}_j = \mathrm{Attention}((\hat{\mathbf{A}}^k\mathbf{X})\mathbf{W}_Q^j, (\hat{\mathbf{A}}^k\mathbf{X})\mathbf{W}_K^j, \mathbf{X}\mathbf{W}_V^j)$ and $\mathbf{W}_O \in \mathbb{R}^{(md_V) \times d_V}$ is a learnable matrix. $\mathbf{Z}^k \in \mathbb{R}^{n \times d_V}$ represents the output of the $k$-th kernel after passing through the multi-head attention block.

\textbf{Adaptive Attention Module.}
With the $k$-th kernel injecting structural information of $k$-hop neighborhood, we can obtain $c + 1$ outputs from $c + 1$ attention kernels, which can be further formulated as $[\mathbf{Z}^0,...,\mathbf{Z}^c]$. Back to the node-level, there are $c + 1$ corresponding output representations for each node. To adaptively capture the structural information encoded by neighborhoods, \name applies a self-attention module to aggregate the corresponding output $[\mathbf{z}_v^0,...,\mathbf{z}_v^c]$ of node $v$. The aggregating process can be written as:
\begin{equation}
    \mathbf{z}_v = \sum_{k=0}^c \alpha_k \mathbf{z}_v^k, 
\end{equation}
where $\mathbf{z}_v$ is the final output of \pname and $\alpha_k$ represents 
the self-attention score.
$\alpha_k$ is calculated by the self-attention mechanism to measure the importance of $\mathbf{z}_v^k$, expressed as:
\begin{equation}
    \alpha_k = \frac{\mathrm{exp} \left (  \sigma(\mathbf{z}_v^k\mathbf{W}) \overrightarrow{\mathbf{w}}^\mathrm{\top} \right )}{\sum_{i=0}^{c} \mathrm{exp} \left (  \sigma(\mathbf{z}_v^i\mathbf{W}) \overrightarrow{\mathbf{w}}^\mathrm{\top}\right )},
\end{equation}
where $\overrightarrow{\mathbf{w}} \in \mathbb{R}^{1 \times d_V}$ is a learnable vector and $\mathbf{W} \in \mathbb{R}^{d_V \times d_V}$ represents a linear projection. By this equation, the calculated attention score $\alpha_k$ signifies the membership strength of $\mathbf{z}_v^k$ in $[\mathbf{z}_v^0,...,\mathbf{z}_v^c]$. Since $\mathbf{z}_v^k$ inherently incorporates the $k$-hop neighborhood information of node $v$, the output $\mathbf{z}_v$ can thus adaptively obtain different hops of neighborhood information.

\subsection{Implementation Details}
Having defined the novel adaptive multi-neighborhood attention, the other components of our adaptive neighborhood-aware transformer follow the original Transformer architecture. Specifically, we replace the first residual term $\mathbf{X}$ with $\mathbf{\hat{A}X}$ to provide complementary local information as \pname can only capture the global information. Consisting of another two linear layers and a GeLU activation, the \name layer can be formally characterized as:
    \begin{align}
    & \mathbf{Z'}^{(l)} = \mathrm{\pname}(\mathrm{LN}(\mathbf{X}^{(l-1)})) + \mathbf{\hat{A}}\mathbf{X}^{(l-1)}, \\
    & \mathbf{X}^{(l)} = \mathrm{FFN}(\mathrm{LN}(\mathbf{Z'}^{(l)})) + \mathbf{Z'}^{(l)},
    \end{align} 
where $\mathrm{LN}$ indicates layer normalization and $l\in \{1,...,L\}$ denotes the $l$-th layer of our \name model.

Besides, for downstream graph property prediction, we need to perform pooling operation to compact node-level representations to a graph-level representation. There are various popular pooling methods, such as by simply taking the sum or average of all node representations, or by adding a virtual node connecting to all nodes in the graph and further training a graph representation.  
 
\subsection{Analysis of the Proposed Method}
Here we give the analysis of our proposed \name as follows:

\textbf{Model Parameters.} 
Comparing to existing graph Transformer models, \name utilizes multiple attention kernels to perform scaled-dot production in parallel. However, existing models usually project one attention kernel to $m=8$ subspaces, \name reduces to $m=2$ or $m=3$ for each attention kernel. Since \name usually sets the number of kernels as $c=3$, the total number of subspaces is roughly equivalent. Besides, \citep{multihead} has verified that the employing three heads for each kernel is sufficient to learn from different positions of the representation. In this way, reducing the number of heads will not weaken the model's expressive capability. 

\textbf{Relations with GNNs.}
The proposed \name can be regarded as a special but powerful graph neural network. Since the \name layer can perform $\mathrm{AGGREGATE}$ and $\mathrm{COMBINE}$ steps of popular GNN models with proper parameters for different attention kernels. With multiple attention kernels based on different neighborhoods, \pname can directly aggregate information from long-range nodes in the graph according the calculated attention score. By replacing the first residual term $\mathbf{X}$ with $\mathbf{\hat{A}X}$, a \name layer can also obtain the local information of the $1$-hop neighborhood. Therefore, the expressive power of \name can go beyond classic message passing GNNs because it can simutaneously capture global and local information.  

\textbf{Relations with existing Graph Transformers.}
In some way, existing graph Transformers such as GraphTrans \cite{graphtrans} and SAT \cite{sat} can be seen as special cases of our proposed \name by choosing proper parameters for specific attention kernels. To generate a model similar to GraphTrans whose attention kernel can be simply expressed as $(\mathrm{GNN}(\mathbf{X}), \mathrm{GNN}(\mathbf{X}), \mathrm{GNN}(\mathbf{X}))$, a multi-layer \name model utilizes a single attention kernel $(\mathbf{\hat{A}}^\alpha \mathbf{X}, \mathbf{\hat{A}}^\alpha \mathbf{X}, \mathbf{\hat{A}}^\alpha \mathbf{X})$ in the first layer and employs a single attention kernel $(\mathbf{X}, \mathbf{X}, \mathbf{X})$ in the following layers. Here $\alpha$ is equal to the number of GNN layers applied in GraphTrans. The attention kernel of the general SAT framework can be written as $(\mathrm{GNN}(\mathbf{X}), \mathrm{GNN}(\mathbf{X}), \mathbf{X})$. Our \name can represent it by leveraging one single attention kernel $(\mathbf{\hat{A}}^\beta \mathbf{X}, \mathbf{\hat{A}}^\beta \mathbf{X}, \mathbf{X})$ in all the layers. Here $\beta$ is equal to the number of GNN layers applied in SAT.    


%% file: 04_Exp.tex
In this section, we evaluate the proposed \name on several graph benchmarks for graph representation learning. Various strong baselines are elaborately selected to compare with \name, such as graph pooling methods, graph neural networks and graph Transformers. To further identify the factors that drives the performance, we conduct ablation studies to analyze the contribution of the components in \name. 

\subsection{Datasets and Baselines}
We first briefly introduce the chosen datasets and baselines here.

\textbf{Datasets.} 
We utilize the following datasets in graph classification tasks to evaluate the performance of \name. NCI1 and NCI109~\cite{nci} are two biochemical datasets where each graph is a compound with atoms and bonds representing the nodes and edges respectively. The label of each graph indicates whether the compound is positive related to anticancer activity. COLLAB~\cite{collab} is a collaboration dataset in which each node is a researcher and each edge represents the collaboration between researchers. The task is to distinguish the research field of each researcher. 
Both ogbg-molpcba and ogbg-code2 are medium datasets for graph classification from the popular leaderboard of Open Graph Benchmark \cite{ogb}.
For ogbg-molpcba, each graph is a molecule and the target is to predict the multiple properties of a molecule. As a computer programming benchmark, ogbg-code2 is a collection of Abstract Syntax Trees (ASTs). By providing the method body of each AST, the task is to predict the sub-tokens forming the method. The statistics of the aforementioned datasets are shown in Table \ref{tab:stat}.

\begin{table}[t]
  \caption{Statistics of the datasets. $\mathrm{V}_{avg}$ ($\mathrm{E}_{avg}$) represents the average number of nodes (edges), respectively.}
  \label{tab:stat}
  \begin{tabular}{lrrrr}
    \toprule
    Dataset & Graphs & $\mathrm{V}_{avg}$ & $\mathrm{E}_{avg}$ & Classes\\
    \midrule
    NCI1 & 4110 & 29.87 & 32.30 & 2\\
    NCI109 & 4127 & 29.68 & 32.13 & 2\\
    COLLAB & 5000 & 74.49 & 2457.78 & 3\\
    ogbg-molpcba & 437929 & 26.00 & 28.10 & 128\\
    ogbg-code2 & 452741 & 125.20 & 124.20 & 5\\
  \bottomrule
\end{tabular}
\end{table}

\textbf{Baselines.} 
To investigate the effectiveness of our proposed \name for graph representation learning, we take three kinds of representative methods as the baselines: 

(\uppercase\expandafter{\romannumeral1}) Recently developed GNN-based methods, including GCN~\cite{gcn}, GAT~\cite{gat} and GraphSage~\cite{graphsage}. GCN is a classic graph neural network that first introduces a hierarchical propagation process to aggregate information from $1$-hop neighborhood layer by layer. GAT applies the attention mechanism from Transformer into graph neural network to distinguish the importance of information from different nodes.  GraphSage provides a general inductive framework that can generate embeddings for unseen nodes by sampling and aggregating features from the local neighborhood of each node. 

(\uppercase\expandafter{\romannumeral2}) Graph pooling methods, including Set2Set~\cite{set2set}, SAGpool~\cite{sagpool} and SortPool~\cite{sortpool}. Set2Set utilizes a recurrent neural network to encode all nodes with content-based attention.
SAGPool is a node-dropping pooling model that selects important nodes according to the self-attention score calculated by GNNs. SortPool directly drops nodes by sorting their representations generated from the previous GNN layers. 

(\uppercase\expandafter{\romannumeral3}) Graph Transformer models, including the original Transformer~\cite{transformer}, GT~\cite{gt}, GraphTrans~\cite{graphtrans} and SAT~\cite{sat}. We have described these models in Section \ref{RW}.  

\subsection{Training Setups}
Here we present the detailed training setups in our experiments.

\textbf{Dataset Split.} We randomly split the dataset into training, validation and test set by a ratio of 8:1:1 for NCI1, NCI109 and COLLAB. For ogbg-molpcba and ogbg-code2, we follow the standard dataset splitting from Open Graph Benchmark \cite{ogb}. 

\textbf{Hyperparameters and Optimization.}
We follow the recommended setting to perform the parameter tuning for all baselines. For the proposed \name, we select the number of \name layers from $\{2, 3, 4, 5\}$, the number of attention kernels from $\{2, 3, 4\}$ and the hidden dimension from $\{128, 256, 300\}$. Besides, we set the learning rate as $2e-4$, the weight decay as $1e-5$, the dropout rate as $0.2$. For all baselines and our proposed \name, we randomly initialize the parameters and optimize with AdamW~\cite{adamw} optimizer with a standard warmup strategy suggested for Transformers in \cite{transformer}. 

\begin{table}
  \caption{Test performance in terms of average accuracy $\pm$ stdev (\%) on 3 benchmarks from TUDatasets. 
  The best results appear in bold. 
  }
  \label{tab:res1}
  \begin{tabular}{lccc}
    \toprule
    Methods & NCI1 & NCI109 & COLLAB\\
    \midrule
    GCN & 79.68 $\pm$ 2.05 & 78.05 $\pm$ 1.59 & 71.92 $\pm$ 3.24 \\
    GAT & 79.88 $\pm$ 0.88 & 79.93 $\pm$ 1.52 & 75.80 $\pm$ 1.60 \\
    GraphSage & 78.98 $\pm$ 1.84 & 77.27 $\pm$ 1.66 & 79.70 $\pm$ 1.70 \\
    \midrule
    Set2Set & 68.62 $\pm$ 1.90 & 69.88 $\pm$ 1.20 & 65.34 $\pm$ 6.44 \\
    SortPool & 73.42 $\pm$ 1.12 & 73.53 $\pm$ 0.91 & 71.18 $\pm$ 2.12 \\
    $\mathrm{SAGPool}_{h}$ & 67.55 $\pm$ 1.03 & 67.91 $\pm$ 0.85 & 73.08 $\pm$ 1.31\\
    \midrule
    Transformer & 68.47 $\pm$ 1.78 & 70.24 $\pm$ 1.29 & 69.57 $\pm$ 3.22\\
    GT & 80.15 $\pm$ 2.04 & 78.94 $\pm$ 1.15 & 79.63 $\pm$ 1.02\\
    GraphTrans & 81.27 $\pm$ 1.90 & 79.20 $\pm$ 2.20 & 79.81 $\pm$ 0.84 \\
    $k$-subtree SAT & 80.69 $\pm$ 1.55 & 79.06 $\pm$ 0.89 & 80.05 $\pm$ 0.55\\
    \midrule
    \name(ours) & \bf{82.73} \textbf{$\pm$} \bf{1.45} & \bf{81.64} \textbf{$\pm$} \bf{2.21} & \bf{81.80} \textbf{$\pm$} \bf{2.24} \\
  \bottomrule
\end{tabular}
\end{table}

\begin{table*}
  \caption{The performance on OGB datasets. The best results appear in bold. }
  \label{tab:res2}
  \begin{tabular}{lcccc}
    \toprule
    &\multicolumn{2}{c}{\textbf{ogbg-molpcba}}& \multicolumn{2}{c}{\textbf{ogbg-code2}}\\
    Methods & Validation AP & Test AP & Validation F1 Score & Test F1 Score \\
    \midrule
    GCN & 0.2059 $\pm$ 0.0033 & 0.2020 $\pm$ 0.0024 & 0.1399 $\pm$ 0.0017 & 0.1507 $\pm$ 0.0018\\
    GCN $+$ Virtual Node & 0.2495 $\pm$ 0.0042 & 0.2424 $\pm$ 0.0034 & 0.1461 $\pm$ 0.0013 & 0.1595 $\pm$ 0.0018 \\
    GIN & 0.2305 $\pm$ 0.0027 & 0.2266 $\pm$ 0.0028 & 0.1376 $\pm$ 0.0016 & 0.1495 $\pm$ 0.0023\\
    GIN $+$ Virtual Node & 0.2798 $\pm$ 0.0025 & 0.2703 $\pm$ 0.0023 & 0.1439 $\pm$ 0.0020 & 0.1581 $\pm$ 0.0026 \\
    \midrule
    Transformer & 0.1316 $\pm$ 0.0012 & 0.1281 $\pm$ 0.0039 & 0.1546 $\pm$ 0.0018 & 0.1670 $\pm$ 0.0015\\
    GraphTrans & 0.2867 $\pm$ 0.0022 & 0.2761 $\pm$ 0.0029 & 0.1661 $\pm$ 0.0012 & 0.1830 $\pm$ 0.0024 \\
    $k$-subtree SAT & - & - & 0.1733 $\pm$ 0.0023 & 0.1937 $\pm$ 0.0028 \\
    \midrule
    \name (ours) & \bf{0.2903} \textbf{$\pm$} \bf{0.0015} & \bf{0.2832} \textbf{$\pm$} \bf{0.0026} & \bf{0.1875} \textbf{$\pm$} \bf{0.0021} & \bf{0.1998} \textbf{$\pm$} \bf{0.0031}\\
    \bottomrule
  \end{tabular}
\end{table*}

\subsection{Performance Comparison}
We compare the performance of our proposed \name with the baseline methods on 5 benchmark datasets for the graph classification task. 

For the three smaller datasets (NCI1, NCI109 and COLLAB) from TUDatasets ~\cite{tudataset}, we train \name for 100 epochs with a batch size of 256. We conduct ten trials with different random seeds for each experiment and calculate the average and standard deviations of the test accuracy. Experimental results are summarized in Table \ref{tab:res1}. Generally speaking, \name consistently outperforms all the baselines, 
demonstrating the superiority of our model. Specifically, \name improves the performance by $1.46\%$, $2.44\%$, $1.75\%$ over the best baselines on NCI1, NCI109 and COLLAB, respectively. 
Comparing with the GNN based methods, the improved performance of \name mainly attributes to its ability of efficiently obtaining long-range structural information meanwhile preserving local structural information. Especially, one can observe that the performance of \name surpasses the graph Transformer counterparts, indicating the effectiveness of our new attention mechanism. Moreover, by adaptively capturing the structural information, our \name can effectively generate more expressive node representations for graph representation learning.

To further validate the superiority of our proposed \name, we conduct experiments on 2 datasets from Open Graph Benchmarks. Since ogbg-molpcba and ogbg-code2 belong to large-scale datasets, we train \name for 30 epochs with a batch size of 16 or 32. By running each experiment 5 times and take the average and standard deviation of the corresponding metric, the obtained experimental results are reported in Table \ref{tab:res2}. Overall, our proposed \name achieves higher performance than other baseline models. Note that 
when comparing to $k$-subtree SAT and GraphTrans that only utilize one attention kernel to extract the structural information, the superior performance of our \name steadfastly confirm the effectiveness of employing multiple attention kernels to adaptively extract the structural information. 

\subsection{Ablation Study}
To analyze the contribution of each component in \name, 
we conduct a series of ablation studies on COLLAB. The experimental results are shown in Table \ref{tab:abl}.

To verify the efficacy of the adaptive attention layer, we replace this component with widely used methods of taking the average, sum, or concatenation of the representations of the nodes. Both average and sum can be regarded as equally aggregating the representations that separately incorporates different local structural information. Simply concatenating the output representations from multiple attention kernels is also a feasible and flexible way to utilize these representations. From Table \ref{tab:abl}, we can observe that the adaptive attention layer significantly improves the performance comparing to other typical methods, further indicating the importance of adaptively capturing the graph structural information.

\subsection{Parameter Study}
In this subsection, we perform a variety of studies to investigate the sensitivity of our proposed \name on three key parameters: the number of propagation steps $k$, the number of heads for single attention kernel $m$, and the number of \name layers $L$. The empirical results on TUDatasets (NCI1, NCI109 and COLLAB) by setting different  parameters are illustrated as follows.

\begin{figure}[]
    \centering
	\includegraphics[]{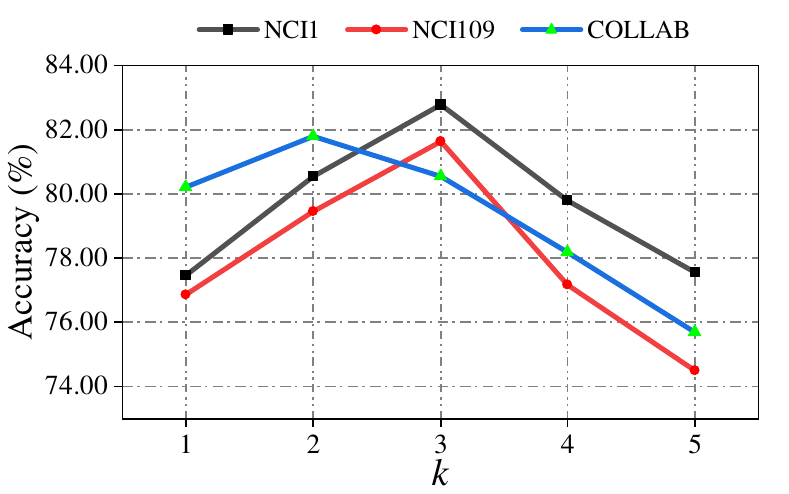}
	\caption{Parameter sensitivity on the propagation step. }
	\label{fig:fig2_c}
\end{figure}

\textbf{On the value of Progation Steps $k$.} 
Since the value of propagation steps $k$ is closely related the number of adpoted attention kernels, we first test the effect on the value of propagation steps $k$ varying from $1$ to $5$ and report the performance of our \name. The results are shown in Figure \ref{fig:fig2_c}. It is notable that with the increment of $k$, the performance significantly raises at first because a larger $k$ brings more affluent structural information. However, the performance starts to decay when the number of kernels exceeds a threshold. The reason is that \name needs suitable number of kernels to encode the structural information. When $k$ is too large, it will inevitably introduce additional redundancies.   

\begin{figure}[]
    \centering
	\includegraphics[]{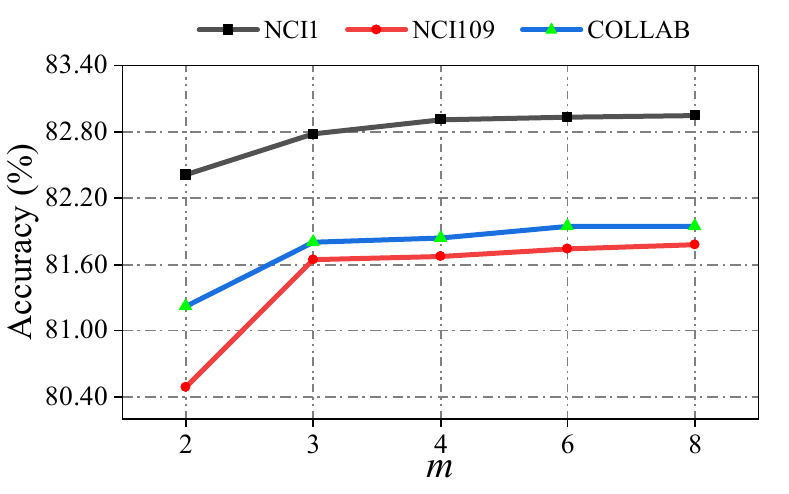}
	\caption{Parameter sensitivity on the number of heads. }
	\label{fig:fig2_m}
\end{figure}

\begin{table}
  \caption{Ablation study on the COLLAB dataset.}
  \label{tab:abl}
  \begin{tabular}{lcc}
    \toprule
    Method & Accuracy & Gains \\
    \midrule
    Sum & 79.68 & $+$ 2.12 \\
    Average & 80.22 & $+$ 1.58\\
    Concatenate & 81.02 & $+$ 0.78\\
    \midrule
    Adaptive attention & 81.80 &  \\
  \bottomrule
\end{tabular}
\end{table}

\begin{figure}[t]
    \centering
	\includegraphics[]{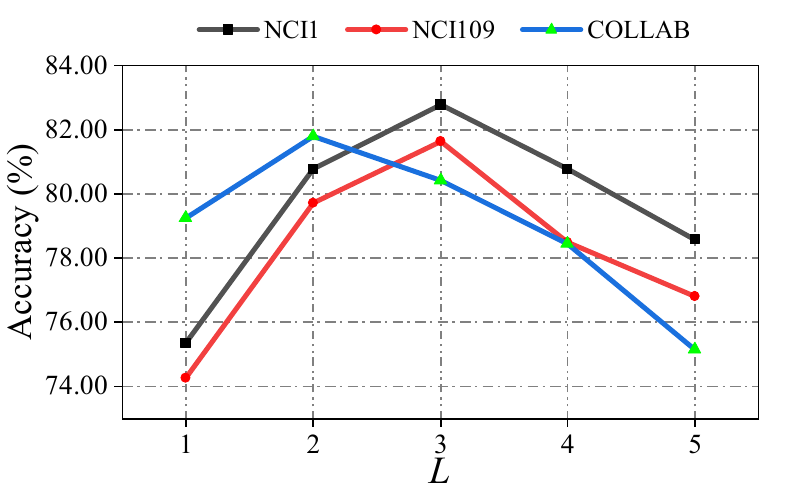}
	\caption{Parameter sensitivity on the number of network layers. }
	\label{fig:fig2_l}
\end{figure}

\textbf{On the Number of Heads $m$.}
To check the impact of the number of heads in a single attention kernel, we further explore the experimental results with a different number of heads. By fixing all other parameters, we vary the number of heads $m$ in $\{2,3,4,6,8\}$ and report the results, as shown in Figure \ref{fig:fig2_m}. Generally speaking, with the increment of the number of heads, the performance of \name continues to increase. Nevertheless, when $m$ lies beyond $3$, there is only a slight performance improvement. The results are also in accord with the analysis in \citep{multihead}. 

\textbf{On the Influence of Layers $L$.}
In some way, the number of layers $L$ is positively relevant to the parameter size of the model. We range $L$ from $1$ to $5$, and the acquired results are reported in Figure \ref{fig:fig2_l}. The results show that \name can achieve higher performance even with a small $L$. By stacking too many layers, the performance starts to degrade as the Transformer based model is easy to suffer from overfitting on small datasets. 

\begin{table}
  \caption{The amount of parameters. 
  In general, \name uses less amount of parameters.
  }
  \label{tab:res_NOP}
  \begin{tabular}{lccc}
    \toprule
    Methods & NCI1 & ogbg-molpcba & ogbg-code2 \\
    \midrule
    GNN & $0.4$M & $3.4$M & $12.5$M \\
    GraphTrans & $0.5$M & $4.2$M & $9.1$M\\
    $k$-subtree SAT & $0.6$M & - & $15.7$M\\
    \name & $0.4$M & $3.9$M & $7.6$M \\
  \bottomrule
\end{tabular}
\end{table}

\subsection{Comparison on the Number of Parameters}
In this subsection, we compare the number of parameters of our proposed \name with GNN, GraphTrans and $k$-subtree SAT. The results are summarized in Table \ref{tab:res_NOP}. As we can see from this Table, \name is significantly more parameter-efficient than GraphTrans and $k$-subtree SAT on the three datasets. The results are reasonable since our proposed \name employs a propagation process that is non-parametric to construct the attention kernels comparing to GraphTrans and $k$-subtree SAT. As for the comparison to GNN, the amount of parameters are roughly at the same scale, but our method gains much better performance.

%% file: 06_Con.tex
In this paper, we propose a novel and powerful graph Transformer called \name for graph representation learning.
Different from existing graph Transformers that capture the structural information by fixed strategies, our \name can adaptively incorporate different structural information based on our new attention module of \pname.
\pname first utilizes different neighborhoods of each node to construct multiple attention kernels with each kernel corresponding to a specific neighborhood.  
By separately performing the self-attention mechanism for each attention kernel, \pname can efficiently generate a series of representations for each node that inject different structural information. 
Then, \pname leverages an attention layer to adaptively aggregate the output representations from different kernels for various nodes.
We further conduct extensive experiments to investigate the effectiveness of \name on five popular datasets. 
Experimental results demonstrate the powerful expressiveness of \name comparing with existing graph Transformers and other mainstream methods.
